\pdfoutput=1

\documentclass[11pt]{article}

\usepackage[preprint]{acl}

\usepackage{times}
\usepackage{latexsym}

\usepackage[T1]{fontenc}

\usepackage[utf8]{inputenc}

\usepackage{microtype}

\usepackage{inconsolata}

\usepackage{graphicx}
\usepackage{amsmath}
\usepackage{enumitem}
\usepackage{booktabs}
\usepackage{multirow}
\usepackage{subcaption}
\usepackage[normalem]{ulem}
\usepackage{fancyvrb}
\usepackage{adjustbox}

\title{Option-ID Based Elimination For Multiple Choice Questions}

\author{
 \textbf{Zhenhao Zhu\textsuperscript{1}},
 \textbf{Bulou Liu\textsuperscript{2}}
  ,
  \textbf{Qingyao Ai\thanks{Corresponding Author}\textsuperscript{2}}, 
  \textbf{Yiqun Liu\textsuperscript{2}}
\\
 \textsuperscript{1}Weiyang College, Tsinghua University
\\
 \textsuperscript{2}Department of Computer Science and Technology, Tsinghua University
 \\\texttt{zhuzhenh22@mails.tsinghua.edu.cn, aiqy@tsinghua.edu.cn} 
}

\begin{document}
\maketitle
\begin{abstract}
Multiple choice questions (MCQs) are a popular and important task for evaluating large language models (LLMs). Based on common strategies people use when answering MCQs, the process of elimination (PoE) has been proposed as an effective problem-solving method. Existing PoE methods typically either have LLMs directly identify incorrect options or score options and replace lower-scoring ones with [MASK]. However, both methods suffer from inapplicability or suboptimal performance.
To address these issues, this paper proposes a novel option-ID based PoE ($\text{PoE}_{\text{ID}}$). $\text{PoE}_{\text{ID}}$ critically incorporates a debiasing technique to counteract LLMs token bias, enhancing robustness over naive ID-based elimination. It features two strategies: $\text{PoE}_{\text{ID}}^{\text{log}}$, which eliminates options whose IDs have log probabilities below the average threshold,  and $\text{PoE}_{\text{ID}}^{\text{seq}}$, which iteratively removes the option with the lowest ID probability. We conduct extensive experiments with 6 different LLMs on 4 diverse datasets. The results demonstrate that $\text{PoE}_{\text{ID}}$, especially $\text{PoE}_{\text{ID}}^{\text{log}}$, significantly improves zero-shot and few-shot MCQs performance, particularly in datasets with more options. Our analyses demonstrate that $\text{PoE}_{\text{ID}}^{\text{log}}$ enhances the LLMs' confidence in selecting the correct option, and the option elimination strategy outperforms methods relying on [MASK] replacement. We further investigate the limitations of LLMs in directly identifying incorrect options, which stem from their inherent deficiencies.
\footnote{Code is avaliable: \url{https://github.com/zzh-thu-22/PoE_ID}}
\end{abstract}

\section{Introduction}

MCQs are a common assessment tool that plays a crucial role in evaluating the reasoning and comprehension abilities of LLMs. Due to their simplicity and structured nature, MCQs are widely used in various contexts such as standardized testing and machine reading comprehension. Therefore, how to improve LLMs' performance on MCQs tasks has become an important area of research.

\begin{figure}[t]
  \includegraphics[width=\columnwidth]{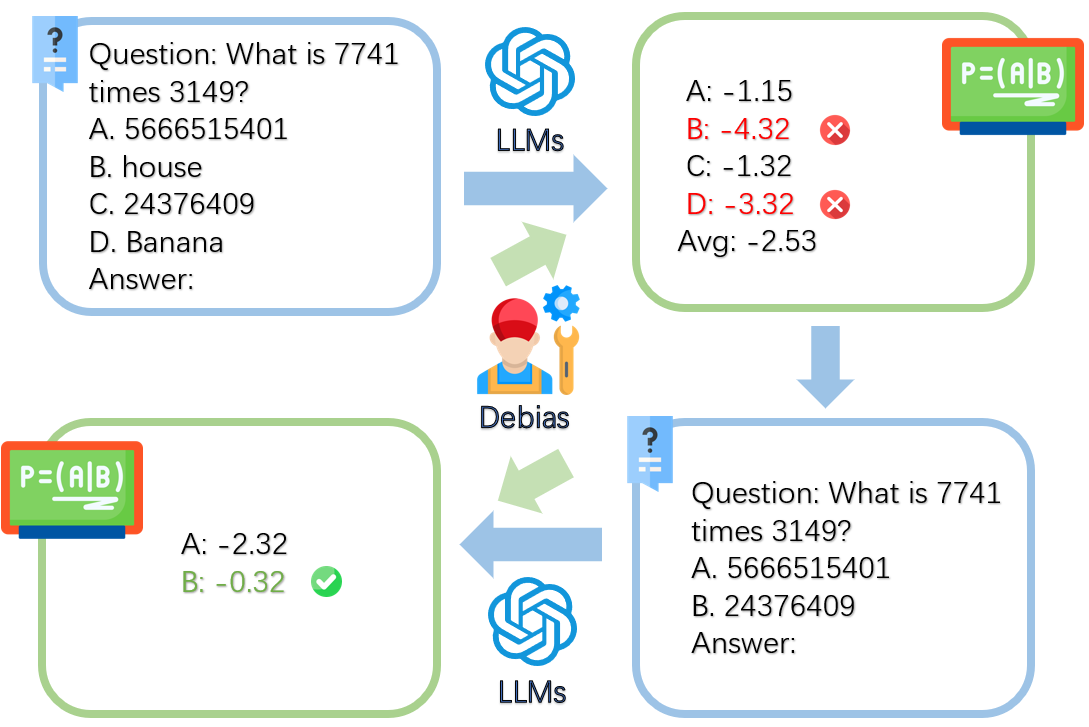}
  \caption{ An illustration of $\text{PoE}_{\text{ID}}^{\text{log}}$ for a multiple choice question. The log probabilities of IDs B and D fall below the average threshold, leading to options B and D being eliminated from the option space.}
  \label{figure1}
\end{figure}

When people solve MCQs, they often use a special strategy named the process of elimination (PoE), which involves first identifying and discarding obviously incorrect options, and then progressively narrowing down the range of possible answers until the correct one is identified. Based on this idea, previous studies have implemented the PoE to LLMs to enhance their MCQs answering capability. Existing PoE for LLMs can be broadly divided into two categories: one involves directly having the LLMs identify and select the incorrect options\footnote{We refer to this method as Explicit Elimination (EE).} \citep{balepur2024easy}, while the other involves scoring each option and replacing those with lower scores with the [MASK] string \citep{ma2023poe}. However, these methods both have significant limitations. First, it has been shown that the PoE based on incorrect options is not applicable to current LLMs. Second, the effectiveness of the PoE based on option scoring is often not as high as that directly answer the MCQs with the option IDs \citep{robinson2023leveraging}.

\begin{table}
\centering
\begin{adjustbox}{width=\columnwidth} 
\begin{tabular}{ccccc}
\toprule

\textbf{Datasets} & \textbf{LLMs} & \textbf{MCP} & \textbf{$\text{PoE}_{\text{ID}}^{\textbf{log}}$}  \\
\midrule
\multirow{2}{*}{Arithmetic\_A} & Gemma-3-12B & 85.4 & \textbf{88.7} \\
 & Qwen2.5-14B & 92.9 & \textbf{98.9} \\

\midrule
\multirow{2}{*}{Arithmetic} & Gemma-3-12B & \textbf{65.0} & 64.8 \\
 & Qwen2.5-14B & \textbf{74.2} & 65.1 \\

\bottomrule
\end{tabular}
\end{adjustbox}
\caption{The answer labels for Arithmetic\_A are uniformly set to A, while the answer labels for Arithmetic are randomly distributed. $\text{PoE}_{\text{ID}}^{\text{log}}$  described here does not include a debiasing step. The introduction of MCP is provided in Section \ref{4.3.2}. The best results are inbold.}
\label{table1}
\end{table}

To address these issues, this study introduces an option-ID based PoE (PoE$_\text{ID}$) alongside two elimination strategies: $\text{PoE}_{\text{ID}}^{\text{log}}$ (Figure \ref{figure1}) and $\text{PoE}_{\text{ID}}^{\text{seq}}$. The former computes the average log-probability of option IDs and eliminates options whose probabilities fall below this threshold. The latter iteratively eliminates the option with the lowest probability. However, experimental results shown in Table \ref{table1}, demonstrate that this method lacks robustness, achieving effectiveness only under specific conditions. Specifically, when the dataset labels are altered from a uniform distribution of all-A labels to a randomized distribution, the method’s performance deteriorates significantly. We attribute this to token bias in current LLMs, where certain option IDs are a priori assigned higher probability mass during MCQs answering, resulting in ID probabilities that fail to accurately reflect the LLMs' true confidence in the options. To mitigate this bias, we introduce a debiasing technique named Pride \citep{zheng2023large}, correcting the raw ID probabilities before proceeding with the elimination process.

 To validate the effectiveness and generalizability of our method, we conduct experiments with 6 different LLMs in zero-shot settings on 4 datasets with varying numbers of options. The experimental results show that PoE$_\text{ID}$ significantly improves the LLMs' performance on MCQs tasks, especially on datasets with a larger number of options. Among the proposed elimination strategies, $\text{PoE}_{\text{ID}}^{\text{log}}$ proves to be the more effective approach. To further validate $\text{PoE}_{\text{ID}}^{\text{log}}$'s effectiveness, our analysis on a subset of datasets and LLMs showed that, post-elimination, the correct option's ID probability substantially increased for most samples. This indicates that our method effectively reduces noise, enabling LLMs to make more accurate judgments. Additionally, we compare two strategies—options elimination versus replacing options with the [MASK] string—and observe that the former significantly outperforms the latter. We also investigate the failure cases of LLMs when directly selecting incorrect options, revealing that these errors primarily arise from the intrinsic limitations of the LLMs themselves. Finally, our findings show that our method maintains robust performance in few-shot settings, further highlighting its generalizability and practical utility.
Specifically, our contributions are threefold:

\begin{itemize}[leftmargin=10pt]

\item We propose a novel option-ID based PoE (PoE$_\text{ID}$), which uniquely incorporates a debiasing technique to counteract token bias in LLMs. This method includes two distinct elimination strategies, $\text{PoE}_{\text{ID}}^{\text{log}}$ and $\text{PoE}_{\text{ID}}^{\text{seq}}$, designed to efficiently refine the answer space in MCQs tasks.

\item We conduct extensive experiments across 6 LLMs and 4 diverse MCQs datasets, demonstrating that PoE$_\text{ID}$, particularly the $\text{PoE}_{\text{ID}}^{\text{log}}$, significantly and robustly improves LLMs' performance in both zero-shot and few-shot settings, especially on datasets with a larger number of options.

\item We provide in-depth analyses that not only elucidate the mechanism by which $\text{PoE}_{\text{ID}}^{\text{log}}$ enhances LLMs reasoning (e.g., by increasing the probability of the correct option's ID post-elimination) but also offer comparative insights into [MASK] replacement and elimination comparison and the intrinsic limitations of LLMs in explicitly identifying incorrect options.

\end{itemize}

\section{Related Work}
\subsection{LLMs}
In recent years, LLMs based on the Transformer \citep{vaswani2017attention} architecture have become the mainstream method. The introduction of models such as the GPT series has accelerated the development of the pre-training-fine-tuning paradigm, demonstrating the remarkable performance of large-scale pre-trained language models across a variety of downstream tasks. Notably, models like GPT-3 \citep{brown2020language} have pushed the boundaries of LLM potential, achieving state-of-the-art performance across multiple tasks. In recent years, with the emergence of studies such as LLaMA (\citealp{touvron2023llama}; \citealp{touvron2023llama2}), researchers have further explored scaling laws and efficient training techniques, enhancing both the performance and efficiency of LLMs. Despite their impressive performance across a wide range of tasks, these LLMs still face challenges, including issues related to model interpretability, bias and fairness \citep{gallegos2024bias}, and their reliance on domain-specific knowledge.

\subsection{MCQs}
MCQs, due to their structured format and ease of evaluation, have become widely adopted in many applications centered around LLMs. For instance, in the automatic evaluation frameworks proposed by (\citealp{chiang2023vicuna}; \citealp{zheng2023judging}), GPT-4 \citep{openai2023gpt4} is provided with a question and two candidate answers and is required to determine which answer is more accurate. This approach leverages the reasoning and comparison capabilities of LLMs to automate the evaluation process, thereby reducing the reliance on human annotations. Additionally, MCQs are commonly used in standard benchmark tests to assess the knowledge and reasoning abilities of language models, such as MMLU \citep{hendrycks2020measuring}. Despite the widespread use of MCQs in evaluation, recent research has raised concerns about their robustness in the context of LLMs. For example, \citet{10.1145/3641289} points out that LLMs exhibit high sensitivity to subtle variations in question phrasing or answer ordering, leading to inconsistent performance across different scenarios. Similarly, \citet{yang2024trust} demonstrates that LLMs often display overconfidence in incorrect answers, particularly when distractors are semantically similar to the correct answer.

\section{Methodology}
\subsection{Problem Setting}
For MCQs, we use \( q \) to denote the question, \( o_i \) to the option ID (e.g. A,B,C,D), \( y_i \) to the option, and \( x \) to the concatenation of the option IDs and the corresponding options.

\subsection{PoE}
For MCQs, two popular methods are (1) directly answer the question and (2) find the correct answer through PoE. The direct answering treats all options equally and relies on the model to directly predict the most likely correct answer. However, this method can be unreliable when the question involves complex relationships between the options and the question, particularly when distractors are closely related to the correct answer. In contrast, PoE is commonly used by people to solve MCQs. This method involves progressively eliminating obviously incorrect options, narrowing down the set of possible answers until the correct one is identified. The application of PoE can significantly enhance the model's ability to handle ambiguous or difficult questions. Given the strong capabilities of LLMs in understanding context, inferring relationships, and reasoning over a large space of potential answers, applying PoE to LLMs offers a promising opportunity to improve performance on MCQs tasks. In contrast to traditional PoE of calculating the probabilities of the options, we use the probabilities of the option IDs as a substitute. After the elimination of options, the option IDs are updated, e.g. when option C is eliminated, the original D will be updated as C. Specially, our method ensures that the relative positions of the remaining options remain unchanged after the elimination of options.

\subsection{Debias}
\citet{zheng2023large} has shown that LLMs exhibit token bias when performing MCQs task, meaning that the LLMs tends to assign more probability mass to certain option ID token a prior when predicting the answer from the option IDs. To mitigate bias, we adopt the Pride method proposed in this research. Pride is a label-free, inference-time debiasing approach that estimates priors using a small set of additional samples, incurring negligible computational overhead. This method enables effective bias reduction while maintaining efficiency and performance.
\[
{P}_{\text{d}}(o_i|q, x) = P(o_i|q, x) / {P}_{\text{prior}}(o_i)
\]

\subsection{Elimination Strategies}

In this section, we introduce two elimination strategies.

\subsubsection{Sequential Elimination ($\text{PoE}_{\text{ID}}^{\textbf{seq}}$)}
\label{3.4.1}
First, we compute the probabilities of the option IDs and select the option corresponding to the ID with the lowest probability as the eliminated option. Then, we remove this option from the original set \( x \), yielding a new set \( x_{\text{new}} \). Using \( x_{\text{new}} \), we recalculate the probabilities for the remaining option IDs. This process is iterated, repeatedly eliminating the option with the lowest probability and updating the set, until only a single option remains.
\[
y_{eli} = \arg \min_{i} {P}_{\text{d}}(o_i|q, x)
\]

\subsubsection{Eliminating Options Below Average Log-Probability ($\text{PoE}_{\text{ID}}^{\textbf{log}}$)}
First, we computes $log {P}_{\text{d}}(o_i|q, x)$, and then calculates an average probability for all options. Options with probabilities below this threshold are eliminated. Finally, we recompute the probabilities for the option IDs based on the \( x_{\text{new}} \) and select the option with the highest ${P}_{\text{d}}(o_i|q, x_{new})$ as the final answer.
\[
y_{eli} = \left\{ y_i \mid \log {P}_{\text{d}}(o_i|q, x) < \text{avg} \right\}
\]

\section{Experimental Setup}
\subsection{Datasets}
We conduct experiments on 4 datasets with varying numbers of options. For each dataset, we randomly sample 1000 instances, prioritizing those from the test set. Detailed data statistics are provided in Appendix \ref{appendixA}.
\begin{itemize}[leftmargin=10pt]

\item \textbf{MMLU-Pro}\citep{wang2024mmlu} represents an advanced extension of the MMLU dataset, designed to rigorously evaluate the reasoning capabilities of LLMs across diverse disciplines, including mathematics, physics, and computer science. Unlike the original MMLU, which primarily assesses general knowledge and commonsense understanding, MMLU-Pro emphasizes complex problem-solving and deep logical reasoning. 

\item \textbf{BIG-Bench}\citep{srivastava2023beyond} We select three representative sub-datasets from the BIG-Bench: Abstract\_Narrative\_Understanding (ABU), Timedial, and Arithmetic. Abstract\_Narrative\_Understanding requires selecting the most relevant proverb given a narrative context, primarily focusing on commonsense reasoning. Timedial involves identifying the correct choice for a masked temporal span within a dialogue context, further emphasizing commonsense inference. In contrast, Arithmetic tasks LLMs with performing arithmetic operations, targeting logical and mathematical reasoning.

\end{itemize}

\subsection{LLM Selection and Configuration}
\begin{itemize}[leftmargin=10pt]
\item \textbf{Selection}
We choose two prominent LLMs families: Qwen and Gemma, covering different parameter scales. Specifically, these include: Qwen2.5-7B/32B \citep{qwen2.5}, Gemma-2-9B(-It) \citep{gemma_2024}, Gemma-3-12B/27B-pt \citep{gemma_2025}. Since We need to obtain output probabilities, all of these LLMs are open-source and available on the HuggingFace platform. 

\item \textbf{Configuration}  
The temperature hyperparameter for all LLMs is set to 0.1. And We use fixed prompts to ensure the consistency of experimental conditions. The specific content of the prompt can be found in Appendix \ref{appendixB}. Unless otherwise specified, all experiments are conducted in zero-shot settings. The experiments are conducted on NVIDIA A100 GPUs with 40GB of memory.

\end{itemize}

\begin{table*}[h]
\centering

\begin{tabular}{ccccccc}
\toprule

\textbf{Datasets} & \textbf{N} & \textbf{LLMs} & \textbf{Max} & \textbf{D-MCP} &\textbf{$\text{PoE}_{\text{ID}}^{\textbf{log}}$} & \textbf{$\text{PoE}_{\text{ID}}^{\textbf{seq}}$} \\
\midrule
\multirow{7}{*}{Timedial} & \multirow{7}{*}{3} & Gemma-2-9B & 78.3 & 79.6 & \textbf{80.4} & 79.8 \\
 &  & Gemma-2-9B-It & 82.6 & 83.0 & 84.1 & \textbf{84.5} \\
 &  & Gemma-3-12B & 81.6 & 83.1 & \textbf{83.2} & 82.3   \\
  &  & Gemma-3-27B  & 86.4 & 87.4 & \textbf{88.5} & 87.8 \\
 &  & Qwen2.5-14B & 67.3 & 83.0 & 84.6 & \textbf{84.9}   \\
 &  & Qwen2.5-32B & 86.0 & \textbf{90.1} & \textbf{90.1} & 89.8   \\
 &  & Average & 80.4 & 84.4 & \textbf{85.2} & 84.9   \\
\midrule
\multirow{7}{*}{Arithmetic} & \multirow{7}{*}{7} & Gemma-2-9B & 50.5 & 50.4 & 57.6 & \textbf{58.7}  \\
&  & Gemma-2-9B-It & 59.8 & 60.2 & 64.0 & \textbf{65.0} \\
 &  & Gemma-3-12B & 65.0 & 64.4 & \textbf{69.4} & 68.9   \\
  &  & Gemma-3-27B  & 65.0 & 77.7 & \textbf{85.3} & 85.1 \\
 &  & Qwen2.5-14B & 84.1 & 79.0 & 85.3 & \textbf{86.0}   \\
 &  & Qwen2.5-32B & 86.1 & 85.8 & \textbf{93.8} & 92.8   \\
 &  & Average & 68.4 & 69.6 & 75.9 & \textbf{76.1}   \\
 
 \midrule
\multirow{7}{*}{Abstract\_Narrative\_Understanding} & \multirow{7}{*}{10} & Gemma-2-9B & 65.1 & 65.6 & 68.7 & \textbf{68.8}  \\
&  & Gemma-2-9B-It & 71.9 &72.4 & \textbf{73.8} & 72.8 \\
 &  & Gemma-3-12B & 63.7 & 63.4 & 66.2 & \textbf{69.2}  \\
  &  & Gemma-3-27B  & 72.9 & 72.3 & \textbf{74.5} & 73.7 \\
 &  & Qwen2.5-14B & 60.9 & 72.2 & \textbf{73.3} & 69.8   \\
 &  & Qwen2.5-32B & 71.5 & 74.2 & 76.3 & \textbf{77.6}   \\
 &  & Average & 67.7 & 70.0 & \textbf{72.1} & 72.0   \\
 
\midrule
 \multirow{7}{*}{MMLU-Pro} & \multirow{7}{*}{10} & Gemma-2-9B & 39.5 & 40.1 & \textbf{41.6} & 40.5  \\
&  & Gemma-2-9B-It & 40.4 & 40.7 & 42.0 & \textbf{43.1} \\
 &  & Gemma-3-12B & 41.2 & 43.0 & \textbf{45.5} & 43.2  \\
  &  & Gemma-3-27B  & 45.5 & 49.3 & 51.3 & \textbf{51.5} \\
 &  & Qwen2.5-14B & 42.8 & 50.0 & \textbf{53.6} & 53.5   \\
 &  & Qwen2.5-32B & 47.5 & 54.6 & \textbf{59.4} & 57.1   \\
 &  & Average & 42.8 & 46.3 & \textbf{48.9} & 48.2   \\
 
\bottomrule
\end{tabular}
\caption{Accuracy (\%) on four datasets. N denotes the number of options. Max denotes the maximum value across all baselines except D-MCP. The best results are inbold. The detailed results can be found in the Appendix \ref{appendixC}.}
\label{table2}
\end{table*}

\subsection{Baselines}
We select 9 common methods as baselines for comparison. These baselines are categorized into four groups.

\subsubsection{Option Scoring Method}
\begin{itemize}[leftmargin=10pt]
\item \textbf{ Language Model (LM)}(baseline in \citealp{zhao2021calibrate}) LM simply selects the option with the highest probability. Thus, LM can be be written as follows:
\[
\arg \max_i P(y_i \mid q)
\]
For causal language models, e.g. GPT, $P(y_i \mid q)$ can be further decomposed as:
\[
P(y_i \mid q) = \prod_{j=1}^{\ell_i} P(y_i^j \mid q, y_i^{1}, \dots, y_i^{j-1})
\]
where $y_i^j$ is the $j$th token of $y_i$ and $\ell_i$ is the number of tokens in $y_i$.

\item \textbf{Average Log Probability (AVG)}\citep{brown2020language} 
AVG takes the logarithm of the probabilities based on the LM and performs normalization.
\[
\arg \max_{i} \frac{1}{\ell_i} \log P(y_i \mid q)
\]

\item \textbf{Domain Conditional PMI ($\text{PMI}_{\text{DC}}$)}\citep{holtzman2021surface} 
$\text{PMI}_{\text{DC}}$ reweights the option score by calculating the probability of the option in the specific task domain, allowing different valid answers to compete fairly.
\[
\arg \max_{i} \log \frac{P(y_i \mid q)}{P(y_i \mid q_{domain})}
\]
Where $q_{domain}$ is the string "Answer:" in our study.

\item \textbf{Channel} \citep{min2022noisy}
Channel computes the probability of the question given the option, assuming that $P(y_i \mid q)$ $\propto$
$P(q \mid y_i)$.
\[
\arg \max_{i} \frac{1}{\ell_i} \log P(q \mid y_i)
\]

\item \textbf{All-Options Log Probability (AOLP)} (baseline in \citealp{ma2023poe})
AOLP calculates the probability of a specific option given all options.
\[
\arg \max_{i}  \log P(y_i \mid q, x)
\]

\end{itemize}

\subsubsection{Option-ID Based Method}

\label{4.3.2}

\begin{itemize}[leftmargin=10pt]
\item \textbf{Multiple Choice Prompt (MCP)}\citep{robinson2023leveraging}
MCP assigns an ID to each option and selects the answer based on the probabilities of IDs.
\[
\arg \max_i P(o_i \mid q,x)
\]
\end{itemize}

\begin{itemize}[leftmargin=10pt]
\item \textbf{Debiased Multiple Choice Prompt (D-MCP)}  
Building on MCP, D-MCP incorporates a debiasing step to mitigate bias in the probability estimates.
\[
\arg \max_i P_{\text{d}}(o_i \mid q, x)
\]
\end{itemize}

\subsubsection{Generative Method}
\begin{itemize}[leftmargin=10pt]
\item \textbf{Text Generation (TG)}(\citealp{wei2022chain}; \citealp{wang2023self}; \citealp{kojima2022zeroshot})
TG first generates a piece of text based on the input, and then extracts the answer from the generated text using manually designed specific rules.
\[
\text{text} = \text{generate}(q, x)
\]
\[
\text{answer} = \text{extract}(\text{text})
\]

\end{itemize}

\subsubsection{PoE Method}
\begin{itemize}[leftmargin=10pt]
\item \textbf{Implicit Elimination (IE)}\citep{ma2023poe}
IE first computes $log P(y_i \mid q, x)$, and then calculates an average probability for all options. Options with probabilities below this threshold are replaced with a special string: [MASK]. Then the probabilities of all options are recalculated, with the probabilities of the replaced options set to negative infinity.
\[
y_{\text{mask}} = \left\{ y_i \mid \log P(y_i \mid q, x) < \text{avg} \right\}
\]
\[
\arg \max_{i} \log P(y_i \mid q, x_{mask})
\]

\end{itemize}

\section{Experimental Results}

\subsection{Comparison with the Baseline}
\label{Section 5.1}
As shown in Table \ref{table2}, D-MCP demonstrates superior overall performance compared to all other baselines. This achievement serves two important implications: (1) it validates the effectiveness of utilizing option IDs for question answering, highlighting LLMs' remarkable symbol-binding capability; (2) it underscores the necessity of our debiasing step, revealing that contemporary LLMs still suffer from inherent biases that require explicit mitigation. Our method achieves consistent state-of-the-art performance across all datasets and LLMs, with both elimination strategies—$\text{PoE}_{\text{ID}}^{\text{seq}}$ and $\text{PoE}_{\text{ID}}^{\text{log}}$—exhibiting comparable performance. However, we identify limitations in the $\text{PoE}_{\text{ID}}^{\text{seq}}$ strategy. Specifically, (1) $\text{PoE}_{\text{ID}}^{\text{seq}}$ incurs higher computational costs due to multiple forward inferences, particularly when the number of options is large, whereas $\text{PoE}_{\text{ID}}^{\text{log}}$ requires only one additional forward inference beyond D-MCP. (2) $\text{PoE}_{\text{ID}}^{\text{seq}}$ is susceptible to error propagation, as early elimination of the correct option renders subsequent steps ineffective, making it challenging to determine the optimal number of options to eliminate. Consequently, we argue that $\text{PoE}_{\text{ID}}^{\text{log}}$ is a more general and robust strategy.
Additionally, we observe that our method’s performance gains on the Timedial dataset, which contains only three options, are notably smaller than on other datasets with more options. We attribute this to reduced noise in datasets with fewer options, enabling LLMs to achieve relatively strong performance even without explicit denoising. The superior performance of D-MCP on Timedial compared to other datasets further corroborates this hypothesis.

\begin{figure}[t]
  \centering

  \includegraphics[width=\columnwidth]{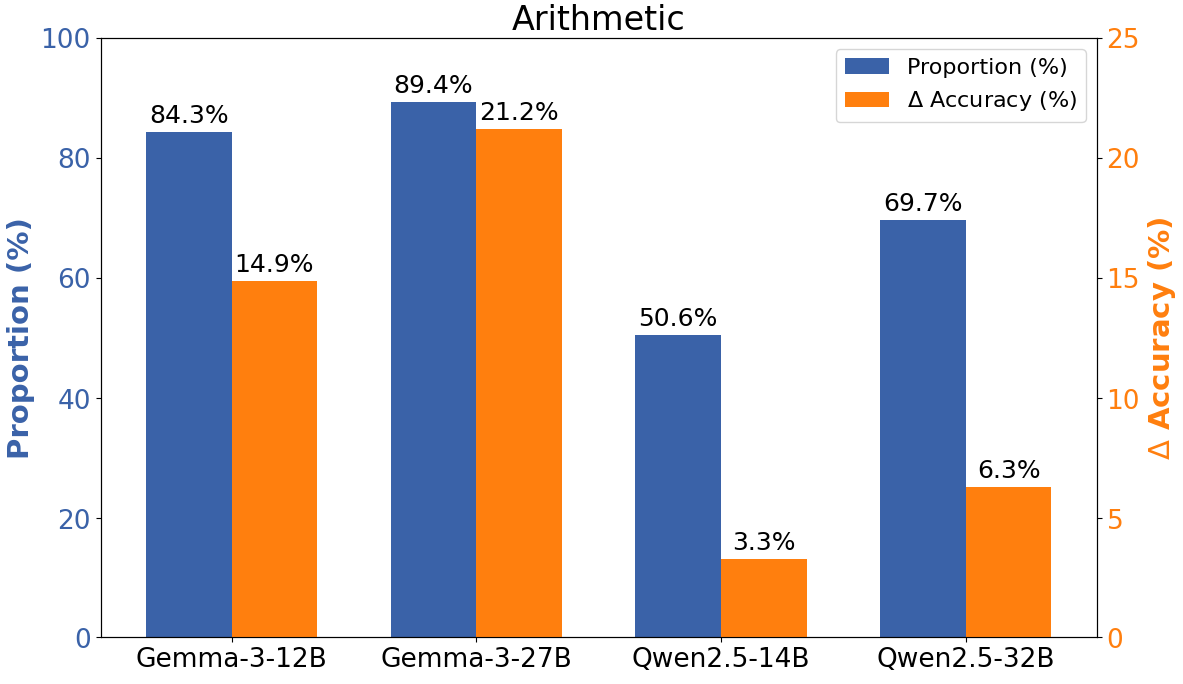} 

  \vspace{0.2cm} 
  
  \includegraphics[width=\columnwidth]{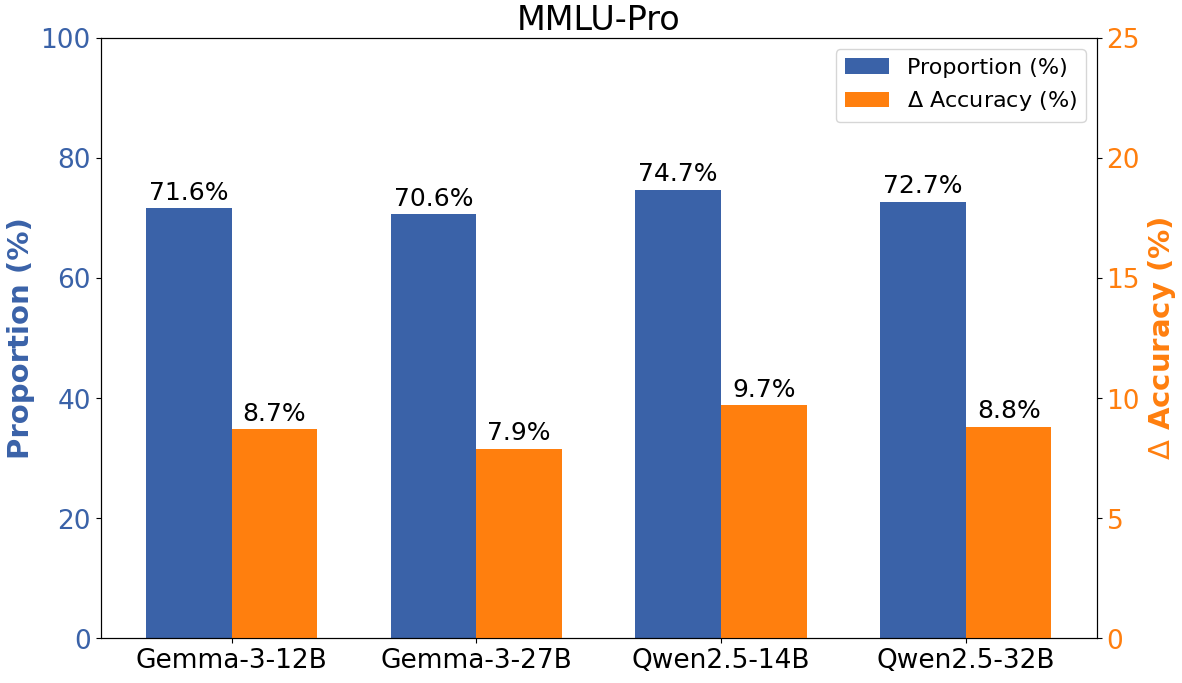}

  \caption{The blue bars represent the proportion of samples where the probability of the correct option's ID increased after the elimination process, while the orange bars show the average magnitude of this probability increase.}
  \label{figure2} 
\end{figure}

\subsection{Analysis of the Effectiveness of $\text{PoE}_{\text{ID}}^{\textbf{log}}$}
After the elimination process, two scenarios emerge: (1) the correct option is no longer in the option space, or (2) the correct option remains in the option space. For the first scenario, while this situation is unavoidable, the elimination is conducted based on D-MCP, ensuring that no samples correctly answered by D-MCP are erroneously eliminated. Consequently, our method does not introduce new errors during the elimination step. Moreover, for samples where D-MCP initially fails, the elimination process retains the correct answer in some cases, providing a second opportunity for accurate inference. For the second scenario, we investigate whether eliminating partial noise enhances the LLMs' confidence in the correct option. To this end, we evaluate two metrics: (1) the proportion of samples where the LLMs' probability for the correct option’s ID increases post-elimination compared to pre-elimination, and (2) the average increase in probability across all samples. We conducted experiments using 4 LLMs—Gemma-3-12B/27B, Qwen2.5-14B/32B—on the Arithmetic and MMLU-Pro datasets to quantify these effects.

Figure \ref{figure2} clearly demonstrates the efficacy of $\text{PoE}_{\text{ID}}^{\text{log}}$. Across both the Arithmetic and MMLU-Pro datasets, a substantial majority of samples (typically over 70\%, and as high as 89.4\% for Gemma-3-27B on Arithmetic) experience an increase in the assigned probability of the correct option's ID after elimination. This prevalent increase signifies effective noise reduction, enabling the LLMs to more reliably identify the correct answer from a refined set of options.
Concurrently, the average magnitude of this probability increase for the correct option ID is consistently positive. For instance, on the Arithmetic dataset, Gemma-3-27B and Gemma-3-12B showed notable average increases of 21.2\% and 14.9\%, respectively. On the MMLU-Pro dataset, all LLMs achieved an average probability uplift of approximately 8-10\% for the correct option.
These results collectively indicate that $\text{PoE}_{\text{ID}}^{\text{log}}$ measurably bolsters the LLMs' confidence in the correct option. Such an increase is crucial for potentially converting previously borderline or incorrect predictions into correct ones, thereby contributing to the method's overall performance enhancement. The consistent positive impact across different LLMs and datasets underscores the effectiveness of our method.

\begin{table}
\centering
\begin{adjustbox}{width=\columnwidth} 
\begin{tabular}{ccccccccc}
\toprule

\textbf{Datasets} & \textbf{LLMs} & \textbf{$\text{IE}_\text{new}$} & $\textbf{MASK}$ & \textbf{$\text{PoE}_{\text{ID}}^{\textbf{log}}$} & \textbf{$\text{PoE}_{\text{ID}}^{\textbf{seq}}$}  \\
\midrule
\multirow{2}{*}{Arithmetic} & G-12B & 62.7 & 59.3 & \textbf{69.4} & 68.9 \\
 & Q-14B & 80.3 & 78.8 & 85.3 & \textbf{86.0}\\

\midrule
\multirow{2}{*}{MMLU-Pro} & G-12B & 39.3 & 39.6 & \textbf{45.5} & 43.2\\
 & Q-14B & 48.2 & 46.1 & \textbf{53.6} & 53.5 \\

\bottomrule
\end{tabular}
\end{adjustbox}
\caption{Accuracy (\%) on two datasets. The best results are inbold.}
\label{table3}
\end{table}

\subsection{[MASK] Replacement vs. Elimination}
Although our method outperforms IE in overall performance, this does not imply that our elimination strategies are necessarily superior to IE's [MASK] replacement strategy. The reason is that IE uses $log P(y_i \mid q, x)$ as the evaluation metric, while we employ $log {P}_{\text{d}}(o_i|q, x)$ for assessment. Based on this, we improve IE and propose a new replacement strategy. Specifically, $\text{IE}_\text{new}$ uses $log {P}_{\text{d}}(o_i|q, x)$ as the evaluation metric.
\[
y_{mask} = \left\{ y_i \mid \log {P}_{\text{d}}(o_i|q, x) < \text{avg} \right\}
\]

The new strategy is based on the strategy mentioned in Section \ref{3.4.1}, which replace option with [MASK] instead of eliminating it. We refer this strategy as MASK.
\[
y_{mask} = \arg \min_{i} {P}_{\text{d}}(o_i|q, x)
\]
We evaluate Gemma-3-12B and Qwen2.5-14B on the Arithmetic and MMLU-Pro datasets.

In the Table~\ref{table3}, we can observe that the two strategies $\text{IE}_\text{new}$ and MASK that replace options with [MASK] string  generally perform worse compared to the $\text{PoE}_{\text{ID}}^{\text{log}}$ and $\text{PoE}_{\text{ID}}^{\text{seq}}$. We attribute this phenomenon to the fact that, despite replacing the options with [MASK] string, the LLMs still tend to assign some probability mass to the IDs corresponding to these options, thereby influencing the probability distribution over the other option IDs. In turn, this affects the LLMs' stability and accuracy.

\begin{table}
\centering
\begin{adjustbox}{width=\columnwidth} 
\begin{tabular}{ccccc}
\toprule

\textbf{Datasets} & \textbf{LLMs} & \textbf{D-MCP} & \textbf{$\text{EE}_{\text{new}}^{\text{log}}$} &  \textbf{$\text{EE}_{\text{new}}^{\text{seq}}$}  \\
\midrule
\multirow{2}{*}{Arithmetic} & G-12B & \textbf{64.4} & 46.5 & 30.9 \\
 & Q-14B & \textbf{79.0} & 64.7 & 40.7\\

\midrule
\multirow{2}{*}{MMLU-Pro} & G-12B & \textbf{43.0} & 26.5 & 12.3 \\
 & Q-14B & \textbf{50.0} & 33.6 & 28.8 \\

\bottomrule
\end{tabular}
\end{adjustbox}
\caption{Accuracy (\%) on two datasets. The best results are inbold.}
\label{table4}
\end{table}

\subsection{The Causes of EE Failure}
In an ideal scenario, the process by which the LLMs select the correct and incorrect option should exhibit consistency. However, the poor performance in EE indicates that the current LLMs still lack sufficient capability in choosing the incorrect options. That being said, we argue that this failure cannot be solely attributed to intrinsic factors of the LLMs. Because different prompts and hyperparameters such as temperature, top-p significantly affect the generated text. So, it is difficult to design a set of optimal answer extraction rules applicable to all scenarios. Based on this, we aim to further investigate the root causes of EE failures, to clarify whether they stem from external factors or from the inherent limitations of the LLMs themselves.
We improve the EE and result in $\text{EE}_{\text{new}}^{\text{log}}$ and $\text{EE}_{\text{new}}^{\text{seq}}$:
\[
\text{Incorrect options} = \left\{ y_i \mid \log {P}_{\text{d}}(o_i|q, x) \geq \text{avg} \right\}
\]
\[
\text{Incorrect option} = \arg \max_i {P}_{\text{d}}(o_i|q, x)
\]
We select Gemma-3-12B, Qwen2.5-14B and conduct experiments on Arithmetic, MMLU-Pro.

In the table~\ref{table4}, We observe a significant decline in performance for both $\text{EE}_{\text{new}}^{\text{log}}$ and $\text{EE}_{\text{new}}^{\text{seq}}$ compared to the D-MCP. By combining the previous research, we speculate that this performance drop is primarily due to intrinsic factors of the LLMs. We believe above phenomenon is closely related to the LLMs' training, which predominantly involves correct answers. As a result, there are fewer samples of incorrect options, limiting the LLMs' ability to understand and identify error options.

\begin{table}
\centering
\begin{adjustbox}{width=\columnwidth} 
\begin{tabular}{ccccccccccc}
\toprule
\multirow{2}{*}{\textbf{Datasets}} & \multirow{2}{*}{\textbf{LLMs}}  & \multicolumn{2}{c}{\textbf{1-shot}} & \multicolumn{2}{c}{\textbf{10-shot}}  \\
\cmidrule(lr){3-4} \cmidrule(lr){5-6} 
 &  & D-MCP & $\text{PoE}_{\text{ID}}^{\text{log}}$ & D-MCP & $\text{PoE}_{\text{ID}}^{\text{log}}$\\
\midrule

\multirow{2}{*}{Arithmetic} & G-12B & 70.9 & \textbf{72.6} & 71.7 &  \textbf{74.1} \\
& Q-14B & 74.7 & \textbf{84.6} & 76.1 & \textbf{81.6}  \\

\midrule

\multirow{2}{*}{ABU} & G-12B & 65.2 & \textbf{66.7} & 66.8 & \textbf{69.5}  \\
& Q-14B & 72.5 & \textbf{73.4} & 72.9 & \textbf{75.0} \\

\midrule

\multirow{2}{*}{MMLU-Pro} & G-12B & 41.2 & \textbf{43.3} & 33.2 & \textbf{45.5} \\
& Q-14B & 52.0 & \textbf{53.7} & 54.4 & \textbf{55.9}  \\

\bottomrule
\end{tabular}
\end{adjustbox}
\caption{Accuracy (\%) on three datasets. The best results are inbold.}
\label{table5}
\end{table}

\subsection{Few-Shot}
To further validate the generality and robustness of our method, we conduct additional experiments under different shot size settings beyond the zero-shot scenario, specifically considering few-shot settings with \( K = 1 \) and \( K = 10 \). For comparison, we select D-MCP. D-MCP, which performs the best under the zero-shot settings, serves as a strong reference for comparison. We evaluate Gemma-3-12B and Qwen2.5-14B on the Arithmetic, Abstract\_Narrative\_Understanding and MMLU-Pro datasets. To ensure the reliability of our findings, we fix the random seed at 0 and draw shots from additional samples used for computing ${P}_{\text{prior}}(o_i)$.

In the Table~\ref{table5}, our experimental results demonstrate that our method continues to outperform the others. 
These results reinforce the effectiveness and robustness of our method, demonstrating that it maintains superior performance even under different shot size conditions.

\section{Conclusion}
This paper introduces $\text{PoE}_{\text{ID}}$, a novel option-ID-based PoE for MCQs tasks, which overcomes limitations of existing methods by incorporating an essential debiasing mechanism. We propose two elimination strategies: $\text{PoE}_{\text{ID}}^{\text{log}}$, which employs a threshold-based approach by eliminating options with log-probabilities below the average, and $\text{PoE}_{\text{ID}}^{\text{seq}}$, which iteratively removes the option with the lowest probability. Extensive experiments conducted across 6 LLMs and 4 datasets demonstrate that $\text{PoE}_{\text{ID}}$, particularly $\text{PoE}_{\text{ID}}^{\text{log}}$, significantly and robustly enhances LLMs' performance in both zero-shot and few-shot settings, with pronounced improvements on questions featuring a larger number of options. Our analysis further reveals that $\text{PoE}_{\text{ID}}^{\text{log}}$ increases confidence in selecting the correct option, outperforms [MASK] replacement strategies through option elimination, and highlights LLMs' limitations in explicitly identifying incorrect options. In summary, $\text{PoE}_{\text{ID}}$ substantially improves LLMs' accuracy on MCQs tasks. Future work will focus on broadening its applicability and further optimizing its components.

\section{Limitations}
Our study has two main limitations. First, we employ fixed prompts without optimizing them, which may constrain the LLMs' performance on specific tasks. Second, we don't explore whether incorporating more complex prompting methods, such as Chain of Thought, could further enhance the LLMs' reasoning abilities and performance. Future research could focus on optimizing these two aspects.

\section*{Acknowledgments}

\bibliography{main}

\appendix

\section{Data Statistics}
\label{appendixA}
\begin{table}[h]
\centering
\begin{adjustbox}{width=\columnwidth} 
\begin{tabular}{ccccccccc}
\toprule

\textbf{Dataset} & \textbf{N} & \textbf{Golden Answer Distribution} \\
\midrule
\multirow{1}{*}{Timedial} &  \multirow{1}{*}{3} & 32.7\% / 33.2\% / 34.1\% \\
 \midrule
\multirow{2}{*}{Arithmetic} &  \multirow{2}{*}{7} & 13.6\% / 14.4\% / 15.8\% \\
 &  & 13.1\% / 13.7\% / 14.2\% / 15.2\%  \\
 \midrule
\multirow{3}{*}{ABU} &  \multirow{3}{*}{10} &  12.1\% / 8.9\% / 8.8\%  \\
 &  & 11.8\% / 9.2\% / 9.5\%  \\
 &  & 10.4\% / 11.5\% / 8.2\% / 9.6\% \\
\midrule
\multirow{3}{*}{MMLU-Pro} & \multirow{3}{*}{10} & 11.3\% / 10.8\% / 9.3\% \\
 &  & 10.4\% / 9.3\% / 10.3\% \\
 &  & 10.2\% / 8.2\% / 10.4\% / 9.8\%  \\
\bottomrule
\end{tabular}
\end{adjustbox}
\caption{Data statistics of datasets used in our experiments. N denotes the number of options.}
\label{table6}
\end{table}

\section{Prompt}
\label{appendixB}
 Examples of the prompts used in our study.
For methods that only require the input of a question, we use the following:
\begin{Verbatim}[fontsize=\small, frame=single]
Question: What is 7741 times 3149?
Answer: 
\end{Verbatim}
For methods that require both a question and options as input, we use the following:
\begin{Verbatim}[fontsize=\small, frame=single]
Question: What is 7741 times 3149?
Choices:
A. 5666515401
B. house
C. 24376409
D. Banana
Answer: 
\end{Verbatim}
For methods that require directly selecting the incorrect options, we use the following:
\begin{Verbatim}[fontsize=\small, frame=single]
Question: What is 7741 times 3149?
Choices:
A. 5666515401
B. house
C. 24376409
D. Banana
Incorrect Option: 
\end{Verbatim}
For few-shot settings, we use the following:
\begin{Verbatim}[fontsize=\small, frame=single]
Question: What is 95 plus 41?
Choices:
A. 136
B. 60140
C. 373
D. house
Answer:A

Question: What is 7741 times 3149?
Choices:
A. 5666515401
B. house
C. 24376409
D. Banana
Answer: 
\end{Verbatim}



\section{Detailed Results}
\label{appendixC}

\begin{table*}
\centering
\begin{adjustbox}{width=\textwidth} 
\begin{tabular}{cccccccccccc}
\toprule

\textbf{LLMs}  & \textbf{LM} & \textbf{AVG} & \textbf{PMI$_{\textbf{DC}}$} & \textbf{Channel} & \textbf{AOLP} & \textbf{IE} & \textbf{TG} & \textbf{MCP} & \textbf{D-MCP} & \textbf{$\text{PoE}_{\text{ID}}^{\textbf{log}}$} & \textbf{$\text{PoE}_{\text{ID}}^{\textbf{seq}}$}\\
\midrule

Gemma-2-9B & 34.8 & 37.8 & 37.5 & 38.7 & 30.3 & 30.3 & 14.1 & 78.3 & 79.6 & 80.4 & 79.8 \\
Gemma-2-9B-It & 59.5 & 60.1 & 59.6 & 30.6 & 57.2 & 58.2 & 80.2 & 82.6 & 83.0 & 84.1 & 84.5  \\
Gemma-3-12B & 36.3 & 37.8 & 42.0 & 28.2 & 28.9 & 28.5 & 29.0 & 81.6 & 83.1 & 83.2 & 82.3  \\
Gemma-3-27B & 36.3 & 39.6 & 40.4 & 33.0 & 30.8 & 30.3 & 34.5 & 86.4 & 87.4 & 88.5 & 87.8  \\
Qwen2.5-14B & 65.8 & 65.1 & 66.8 & 36.2 & 63.3 & 61.8 & 11.2 & 67.3 & 83.0 & 84.6 & 84.9  \\
Qwen2.5-32B & 71.4 & 70.2 & 72.7 & 43.3 & 64.7 & 64.9 & 26.9 & 86.0 & 90.1 & 90.1 & 89.8  \\

\bottomrule
\end{tabular}
\end{adjustbox}
\caption{Accuracy (\%) on  Timedial.}
\end{table*}

\begin{table*}
\centering
\begin{adjustbox}{width=\textwidth} 
\begin{tabular}{cccccccccccc}
\toprule

\textbf{LLMs}  & \textbf{LM} & \textbf{AVG} & \textbf{PMI$_{\textbf{DC}}$} & \textbf{Channel} & \textbf{AOLP} & \textbf{IE} & \textbf{TG} & \textbf{MCP} & \textbf{D-MCP} & \textbf{$\text{PoE}_{\text{ID}}^{\textbf{log}}$} & \textbf{$\text{PoE}_{\text{ID}}^{\textbf{seq}}$}\\
\midrule

Gemma-2-9B & 3.7 & 18.0 & 10.2 & 18.3 & 13.1 & 13.2 & 9.8 & 50.5 & 50.4 & 57.6 & 58.7\\
Gemma-2-9B-It & 18.7 & 46.3 & 40.5 & 21.1 & 21.6 & 21.6 & 58.1 & 59.8 & 60.2 & 64.0 & 65.0 \\
Gemma-3-12B & 2.7 & 18.7 & 11.1 & 25.4 & 15.9 & 14.4 & 1.7 & 65.0 & 64.4 & 69.4 & 68.9 \\
Gemma-3-27B & 2.7 & 19.7 & 14.5 & 9.8 & 17.8 & 19.5 & 1.0 & 65.0 & 77.7 & 85.3 & 85.1 \\
Qwen2.5-14B & 79.3 & 81.9 & 84.1 & 18.5 & 73.2 & 77.3 & 2.0 & 74.2 & 79.0 & 85.3 & 86.0  \\
Qwen2.5-32B & 81.0 & 84.1 & 84.4 & 19.5 & 75.4 & 77.4 & 0.0 & 86.1 & 85.8 & 93.8 & 92.8 \\

\bottomrule
\end{tabular}
\end{adjustbox}
\caption{Accuracy (\%) on Arithmetic.}
\end{table*}

\begin{table*}
\centering
\begin{adjustbox}{width=\textwidth} 
\begin{tabular}{cccccccccccc}
\toprule

\textbf{LLMs}  & \textbf{LM} & \textbf{AVG} & \textbf{PMI$_{\textbf{DC}}$} & \textbf{Channel} & \textbf{AOLP} & \textbf{IE} & \textbf{TG} & \textbf{MCP} & \textbf{D-MCP} & \textbf{$\text{PoE}_{\text{ID}}^{\textbf{log}}$} & \textbf{$\text{PoE}_{\text{ID}}^{\textbf{seq}}$}\\
\midrule

Gemma-2-9B & 10.6 & 11.0 & 14.5 & 10.7 & 10.8 & 9.6 & 43.6 & 65.1 & 65.6 & 68.7 & 68.8 \\
Gemma-2-9B-It & 23.8 & 20.2 & 13.2 & 9.9 & 31.9 & 32.3 & 51.1 & 71.9 & 72.4 & 73.8 & 72.8  \\
Gemma-3-12B & 12.2 & 10.4 & 13.7 & 10.9 & 12.1 & 9.2 & 40.2 & 63.7 & 63.4 & 66.2 & 69.2  \\
Gemma-3-27B & 12.9 & 11.1 & 10.4 & 9.5 & 10.6 & 9.8 & 51.6 & 72.9 & 72.3 & 74.5 & 73.7  \\
Qwen2.5-14B & 20.6 & 14.9 & 24.1 & 35.5 & 22.8 & 21.9 & 11.8 & 60.9 & 72.2 & 73.3 & 69.8  \\
Qwen2.5-32B & 21.0 & 14.1 & 19.9 & 34.4 & 30.7 & 30.4 & 27.4 & 71.5 & 74.2 & 76.3 & 77.6  \\

\bottomrule
\end{tabular}
\end{adjustbox}
\caption{Accuracy (\%) on Abstract\_Narrative\_Understanding.}
\end{table*}

\begin{table*}
\centering
\begin{adjustbox}{width=\textwidth} 
\begin{tabular}{cccccccccccc}
\toprule

\textbf{LLMs}  & \textbf{LM} & \textbf{AVG} & \textbf{PMI$_{\textbf{DC}}$} & \textbf{Channel} & \textbf{AOLP} & \textbf{IE} & \textbf{TG} & \textbf{MCP} & \textbf{D-MCP} & \textbf{$\text{PoE}_{\text{ID}}^{\textbf{log}}$} & \textbf{$\text{PoE}_{\text{ID}}^{\textbf{seq}}$}\\
\midrule

Gemma-2-9B & 8.4 & 9.3 & 12.8 & 12.5 & 12.4 & 11.3 & 22.1 & 39.5 & 40.1 & 41.6 & 40.5 \\
Gemma-2-9B-It & 13.3 & 15.5 & 17.9 & 12.5 & 21.7 & 22.4 & 29.0 & 40.4 & 40.7 & 42.0 & 43.1 \\
Gemma-3-12B & 7.2 & 8.6 & 10.7 & 13.2 & 11.1 & 10.3 & 18.7 & 41.2 & 43.0 & 45.5 & 43.2 \\
Gemma-3-27B & 7.2 & 8.3 & 13.1 & 12.4 & 10.2 & 10.8 & 28.6 & 45.5 & 49.3 & 51.3 & 51.5  \\
Qwen2.5-14B & 24.5 & 26.9 & 28.9 & 17.7 & 36.0 & 35.8 & 24.3 & 42.8 & 50.0 & 53.6 & 53.5 \\
Qwen2.5-32B & 27.6 & 30.4 & 31.8 & 17.3 & 40.0 & 38.1 & 22.5 & 47.5 & 54.6 & 59.4 & 57.1 \\

\bottomrule
\end{tabular}
\end{adjustbox}
\caption{Accuracy (\%) on MMLU-Pro.}
\end{table*}

\onecolumn

\section{LLMs and Datasets}
\begin{table}[!htbp] 
\centering
\begin{tabular}{ccccccccc}
\toprule

\textbf{LLMs} & \textbf{URL} \\
\midrule
Gemma-2-9B & \url{https://huggingface.co/google/gemma-2-9b}  \\
Gemma-2-9B-It & \url{https://huggingface.co/google/gemma-2-9b-it}  \\
Gemma-3-12B & \url{https://huggingface.co/google/gemma-3-12b-pt}  \\
Gemma-3-27B & \url{https://huggingface.co/google/gemma-3-27b-pt}  \\
Qwen2.5-14B & \url{https://huggingface.co/Qwen/Qwen2.5-14B}  \\
Qwen2.5-32B & \url{https://huggingface.co/Qwen/Qwen2.5-32B}  \\

\bottomrule
\end{tabular}
\end{table}

\begin{table}[!htbp] 
\centering
\begin{tabular}{ccccccccc}
\toprule

\textbf{Datasets} & \textbf{URL} \\
\midrule
MMLU-Pro & \url{https://huggingface.co/datasets/TIGER-Lab/MMLU-Pro}  \\
BIG-Bench & \url{https://huggingface.co/datasets/hails/bigbench}  \\

\bottomrule
\end{tabular}
\end{table}

\end{document}